\documentclass[sigconf]{acmart}
\usepackage{tabularx}
\AtBeginDocument{%
  }

\setcopyright{acmlicensed}
\copyrightyear{2018}
\acmYear{2018}
\acmDOI{XXXXXXX.XXXXXXX}
\acmConference[Conference acronym 'XX]{Make sure to enter the correct
  conference title from your rights confirmation email}{June 03--05,
  2018}{Woodstock, NY}
\acmISBN{978-1-4503-XXXX-X/2018/06}

\begin{document}

\title{SIA: A Synthesize-Inject-Align Framework for Knowledge-Grounded and Secure E-commerce Search LLMs with Industrial Deployment}

\author{Zhouwei Zhai}
\authornote{The corresponding author.}
\orcid{1234-5678-9012}
\email{zhaizhouwei1@jd.com}
\affiliation{%
  \institution{JD.com}
  \city{Beijing}
  \country{China}
}

\author{Mengxiang Chen}
\email{chenmengxiang9@jd.com}
\affiliation{%
  \institution{JD.com}
  \city{Beijing}
  \country{China}
}

\author{Anmeng Zhang}
\email{zhanganmeng1@jd.com}
\affiliation{%
  \institution{JD.com}
  \city{Beijing}
  \country{China}
}
\renewcommand{\shortauthors}{Zhai et al.}

\begin{abstract}
Large language models offer transformative potential for e‑commerce search by enabling intent‑aware recommendations. However, their industrial deployment is hindered by two critical challenges: (1) knowledge hallucination due to insufficient encoding of dynamic, fine‑grained product knowledge, and (2) security vulnerabilities under jailbreak attacks that threaten compliance. To address these issues, we propose SIA—a Synthesize‑Inject‑Align framework for building knowledgeable and secure e‑commerce search LLMs. Our approach first synthesizes high‑quality natural language corpus by combining structured knowledge graphs with unstructured behavioral logs, augmented with reasoning chains and safety‑aware data. We then introduce a parameter‑efficient pre‑training strategy based on Depth Up‑Scaling to inject domain knowledge while preserving general capabilities. Finally, a dual‑path alignment method via multi‑task instruction tuning and adversarial training strengthens both task performance and safety robustness. The framework has been deployed at JD.com,China's largest self‑operated e‑commerce platform, where A/B tests across five core search scenarios demonstrate significant improvements in key business metrics, validating its industrial effectiveness and scalability.
\end{abstract}


\maketitle

\section{Introduction}
As a core component of the digital economy, e-commerce search systems directly determine user experience and platform efficiency.  LLMs\cite{brown2020language,achiam2023gpt,touvron2023llama,guo2025deepseek} present an opportunity to shift the paradigm of e-commerce search from "keyword matching" toward "intent understanding + reasoning-based guidance." However, the large-scale deployment of general-purpose LLMs in e-commerce faces two major challenges: first,e-commerce knowledge hallucination—general models lack deep encoding of dynamic and fine-grained e-commerce knowledge (e.g., SKU relations, user behaviors), leading to factual errors; second,security and compliance risks—general safety guardrails can be easily bypassed by jailbreak prompts (e.g., inducing recommendations of prohibited goods), posing legal and risk-control hazards.
For knowledge injection, existing studies enhance knowledge density through continued pre-training on domain corpora\cite{gururangan2020don} and fine-tuning (e.g., EcomGPT\cite{ma2023ecomgpt,li2024ecomgpt}, eCeLLM\cite{peng2024ecellm}), but these often trigger catastrophic forgetting. Parameter-efficient fine-tuning  methods such as LoRA and Adapter\cite{hu2022lora,houlsby2019parameter,dettmers2023qlora,zhang2023adalora} reduce costs, yet their limited parameter capacity struggles to encode complex knowledge\cite{biderman2024lora}. 

Retrieval-augmented generation (RAG)\cite{lewis2020retrieval} relies on retrieval accuracy and introduces additional latency. In terms of application, many LLM-enhanced solutions employ distilled smaller models, which compromises reasoning capability\cite{hinton2015distilling,magister2023teaching}; generative retrieval methods\cite{tay2022transformer,bevilacqua2022autoregressive} directly generate product IDs but are constrained by catalog scale and update frequency\cite{pradeep2023how}. For safety alignment, existing approaches (e.g., RLHF\cite{ouyang2022training}, DPO\cite{rafailov2023direct}) primarily rely on general safety datasets and lack coverage of e-commerce-specific risks.

The core difficulty lies in: how to inject large-scale, heterogeneous, and dynamic e-commerce knowledge without sacrificing general reasoning ability, while establishing robust safety guardrails? To address this, we propose the SIA (Synthesize, Inject, Align) framework, an end-to-end solution encompassing "data synthesis → knowledge injection → domain alignment" to develop an e-commerce search LLM that is both knowledge-grounded and safety-compliant. Our main contributions are as follows:
\begin{itemize}
    \item \textbf{Novel e-commerce knowledge data synthesis method}: Combining knowledge graphs and behavioral logs, we leverage LLMs to generate high-quality natural language corpora, synthesize explicit chain-of-thought data to enhance reasoning, and construct safety-aware pre-training data to mitigate knowledge sparsity and security vulnerabilities at the source.
    \item \textbf{Efficient knowledge injection pre-training technique}: We introduce a parameter expansion approach based on Depth Up-Scaling\cite{kim2023solar} coupled with layer initialization and learning strategies, to inject massive e-commerce knowledge and safety corpora while preserving general capabilities and alleviating forgetting.
    \item \textbf{Fine-grained domain-enhanced alignment method}: We design a dual-path alignment strategy combining multi-task instruction tuning on the task side and adversarial training on the safety side, along with a balanced mix of general alignment data, to strengthen model performance on e-commerce subtasks and robustness in high-sensitivity dimensions.
    \item \textbf{Industrial-scale validation and deployment}: Our method has been deployed on the JD.com platform. A/B testing across scenarios including search suggestion, product title generation, review summarization, query correction, and safety moderation demonstrates significant improvement in key metrics, proving its industrial applicability and reusability.
    \end{itemize}
\section{Related Work}
\subsection{Domain Adaptation of Large Language Models}
Adapting general-purpose large language models to vertical domains is a prominent research focus. Mainstream approaches, such as Continued Pre-training (CPT), offer the most direct means to inject domain knowledge into models, significantly reducing the perplexity of domain-specific texts\cite{gururangan2020don}. However, CPT is often prone to the risk of catastrophic forgetting. Although mixing in general-domain data replay\cite{gupta2023continual} is a common mitigation strategy, balancing domain depth with general breadth remains challenging at an industrial scale. Techniques represented by LoRA\cite{hu2022lora} and Adapter\cite{houlsby2019parameter} achieve adaptation by freezing most parameters and fine-tuning only a small set of incremental parameters, yet they struggle with tasks requiring deep integration into complex knowledge systems, such as e-commerce SKU graphs. EcomGPT\cite{ma2023ecomgpt,li2024ecomgpt} and eCeLLM\cite{peng2024ecellm} enhance performance on e-commerce tasks through domain-specific CPT and supervised fine-tuning, but do not adequately address the retention of general capabilities or domain-specific safety considerations.RAG\cite{lewis2020retrieval} introduces external knowledge via retrieval without updating model parameters, thereby reducing hallucinations. However, its effectiveness is constrained by retrieval accuracy, and long-context retrieval can easily introduce noise that degrades generation quality.

\subsection{LLMs in E-Commerce Search}

E-commerce search has evolved from traditional keyword matching techniques, such as TF-IDF\cite{salton1988term}, to semantic matching\cite{huang2020embedding,reimers2019sentence}. This evolution has entered a new phase characterized by large language model (LLM) augmentation, generative retrieval, and conversational shopping guides, driven by the rise of LLMs. LLM augmentation primarily leverages LLMs for query expansion\cite{nguyen2025minielm,dai2024enhancing,peng2024large}, intent classification\cite{yuan2025semi}, or relevance judgment\cite{tang2025lref,dong2025taosr1}. However, due to the stringent industrial requirement for millisecond-level inference latency, only distilled small models can typically be deployed, compromising their complex reasoning capabilities.Generative retrieval attempts to directly generate document identifiers\cite{tay2022transformer,bevilacqua2022autoregressive,li2024generative}, aiming to replace traditional index-and-retrieval pipelines. Yet, its scalability and real-time performance face significant challenges within e-commerce product catalogs characterized by billions of listings and high-frequency updates.Conversational shopping guides utilize APIs such as ChatGPT to enable natural interaction\cite{gao2023chat}. However, lacking deep awareness of platform-specific product attributes, promotion rules, and inventory status, they are prone to generating “e-commerce hallucinations.”
\subsection{LLM Safety and Alignment}
Ensuring the safety and compliance of LLMs primarily relies on two categories of methods: RLHF\cite{ouyang2022training} and red teaming. RLHF, as the dominant alignment paradigm, commonly employs the PPO algorithm\cite{schulman2017proximal} to optimize for human preferences, while DPO\cite{rafailov2023direct} is gaining popularity due to its more stable training. Red teaming probes model defense boundaries via automated adversarial attacks (e.g., GCG\cite{zou2023universal}, AutoDAN\cite{liu2023jailbreaking}). However, its datasets predominantly focus on risks within generic domains. Safety challenges in e-commerce exhibit high specificity and covertness. Generic safety safeguards struggle to effectively identify its unique violation patterns, and simple, general-purpose alignment proves insufficient to adequately address these long-tail, high-risk, domain-specific scenarios.
\section{The SIA Framework: Overview}

\begin{figure}[h]
  \centering
  \includegraphics[width=\linewidth]{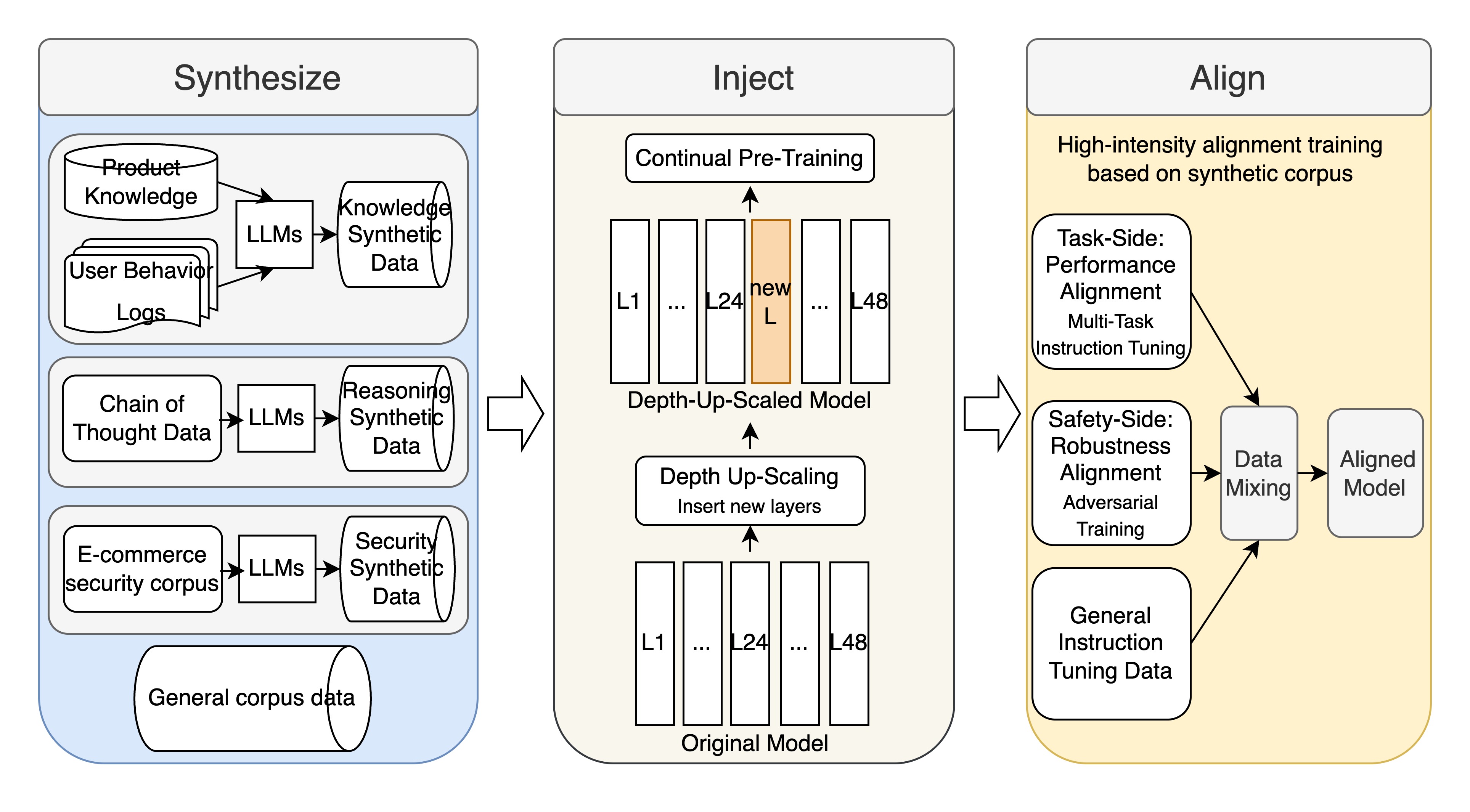}
  \caption{SIA Overview}
  \label{figmain}
\end{figure}

The SIA framework (Figure \ref{figmain}) is built upon the core philosophy of achieving a dynamic equilibrium between specialization/generality and intelligence/security through data-algorithm synergy.

Our framework consists of three stages:
\begin{itemize}
    \item \textbf{Synthesize: Multi-source E-commerce Knowledge Data Synthesis}
    
    This stage transforms heterogeneous platform data into high-quality natural language “textbooks.” Specifically, it includes:
    \begin{itemize}
        \item \textbf{Contextual Synthesis of Knowledge Graphs and Behavioral Logs}: Leveraging LLMs as “data transformers” to synthesize structured knowledge and unstructured logs into semantically rich natural language descriptions, enhancing knowledge absorption efficiency in subsequent pre-training.
        \item \textbf{Chain-of-Thought Reasoning Data Synthesis}: Generating data containing explicit reasoning paths to augment the model’s multi-step reasoning capability.
        \item \textbf{E-commerce Safety Corpus Enhancement}: Systematically defining high-risk e-commerce topics and synthesizing targeted safety Q\&A and adversarial instruction data to build domain-specific safety safeguards from the outset.
    \end{itemize}
    \item \textbf{Inject: Knowledge Injection Pre-training via Parameter Expansion}
    
To efficiently inject knowledge while avoiding catastrophic forgetting, we employ depth up-scaling-based parameter expansion: Increasing the number of Transformer layers within the original model, combined with carefully designed layer initialization strategies and dynamic learning rate scheduling. This strategy guides the newly added capacity to prioritize absorbing e-commerce and safety knowledge, achieving an optimal balance between domain capability enhancement and general capability preservation.
    \item \textbf{Align: Dual-Path Domain-Enhanced Alignment}
    
This stage precisely directs model capabilities towards downstream tasks via refined data strategies:
    \begin{itemize}
        \item \textbf{Task-Side Performance Alignment}: Performing instruction fine-tuning by adjusting the granularity and mixing ratios of multi-task instructions to optimize task performance.
        \item \textbf{Safety-Side Robustness Alignment}: Conducting high-intensity systematic adversarial training based on the safety corpus synthesized in Stage 1, and combining it with generic alignment data via proportioned blending to comprehensively enhance the model’s safety and compliance.
    \end{itemize}
\end{itemize}

\section{Synthesize: Knowledge-Aware Data Synthesis}

\subsection{Context-Aware Knowledge Transformation}
We propose a context-aware method for generating product knowledge data, whose core innovation lies in “translating” multi-source heterogeneous e-commerce data into information-dense, fluent, and natural textual descriptions, thereby significantly enhancing the model’s efficiency in absorbing e-commerce entities and their complex relationships. This method consists of two key steps:
\begin{enumerate}
    \item \textbf{Multi-Source Heterogeneous Information Aggregation}: A unified data view centered around products is constructed, aggregating contextual information from diverse sources:
    \begin{itemize}
        \item \textbf{Structured Knowledge}: Core entities and relationships are extracted from the product knowledge graph, including precise category hierarchical paths, brand, material, dimensions, specifications, and other attribute key-value pairs.
        \item \textbf{Semi-Structured Text}: The aggregation integrates seller-provided detailed descriptions, as well as filtered informative product reviews and Q\&A content.
    \end{itemize}

    \item \textbf{Prompt Engineering-Driven Multi-Task Text Generation}: Leveraging the aggregated information, a powerful general-purpose LLM (utilizing Deepseek-R1\cite{guo2025deepseek}) is employed for synthesis. We guide the LLM to generate diverse forms of high-quality textual data through four types of tasks:
    \begin{itemize}
        \item \textbf{Product Description}: Given product knowledge information, the LLM synthesizes objective, accurate, and comprehensive product descriptions that integrate all key attributes.
        \item \textbf{Attribute Q\&A}: Based on the input product knowledge, the LLM generates question-answer pairs targeting different attributes.
        \item \textbf{Product Comparison}: Using information from two input products, the LLM synthesizes comparative descriptions analyzing the products across various dimensions.
        \item \textbf{Review Summarization}: Provided with product information and review records, the LLM produces objective summaries categorized into positive, negative, and neutral aspects.
    \end{itemize}

\end{enumerate}
\subsection{Chain-of-Thought Synthesis for Search Reasoning}
To improve the model's implicit reasoning in multi-step e-commerce search, we use real user logs and Deepseek-R1\cite{guo2025deepseek} to infer latent decision logic, synthesizing instruction data with explicit reasoning chains.
The specific synthesis workflow is as follows:
\begin{enumerate}
    \item \textbf{Real Session Log Mining}: 
    \begin{itemize}
        \item \textbf{Query Error Correction} (user-initiated corrections or system auto-corrections).
        \item \textbf{Click-Purchase Behavior Sequence} (the complete pathway from query and impression to click and purchase).
    \end{itemize}
    \item \textbf{Chain-of-Thought Reasoning Synthesis}:For the above instances, generate corresponding “Reasoning-Answer” pairs based on the LLM, explicitly restoring their decision-making thought chains. This includes 2 task types:
    \begin{itemize}
        \item \textbf{Error Correction Reasoning}: Input: Validated erroneous or reformulated query pairs. Output: The LLM analyzes the intent and then synthesizes a reasoning chain, including reasoning steps, and the final corrected answer.
        \item \textbf{Shopping Decision Reasoning}: Input: User query, list of product results, user click behavior, final purchased product. Output: The LLM analyzes and synthesizes the reasoning process underlying the user’s purchase decision.
    \end{itemize}
\end{enumerate}

\subsection{Safety-Centric Data for Proactive Risk Mitigation}
\begin{figure}[h]
  \centering
  \includegraphics[width=\linewidth]{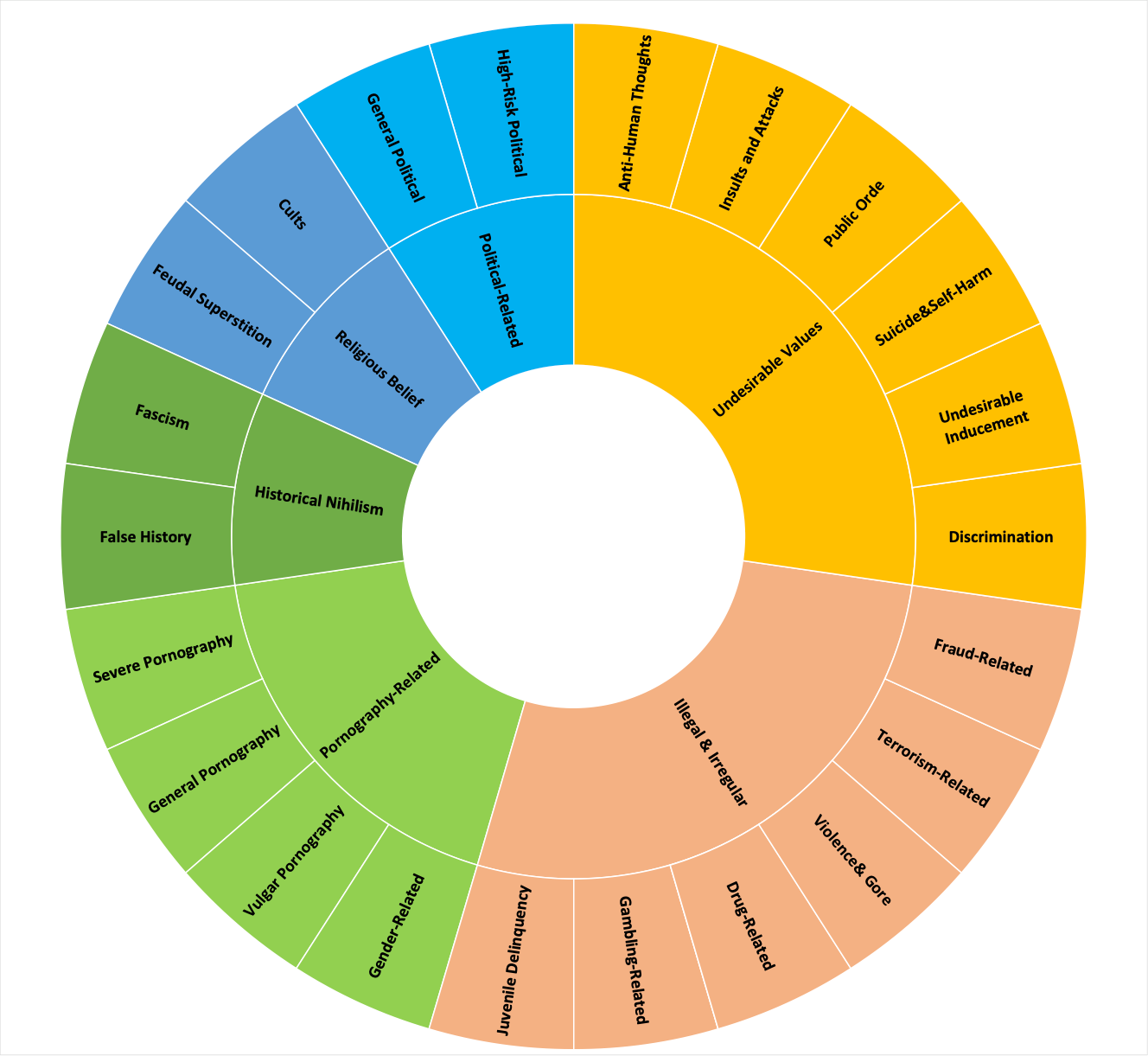}
  \caption{safety scenario}
  \label{fig:safety}
\end{figure}

During the source phase of data synthesis, we systematically construct comprehensive safety‑aligned corpora, aiming to instill intrinsic safety awareness in models from the ground up.We define a high‑risk topic framework comprising six top‑level categories spanning the entire e‑commerce ecosystem, including: Political-Related, Undesirable Values, Illegal and Irregular, Pornography-Related, Historical Nihilism, Religious Belief as in Figure \ref{fig:safety}.

For each category of topics, we synthesized a trifecta of defensive data using an LLM (Deepseek-R1\cite{guo2025deepseek}):
\begin{enumerate}
    \item \textbf{Safe QA Pairs}: Directly teaching the model the standard paradigm for compliant responses. For example:
    \begin{itemize}
        \item Question: "How can I buy [a certain prohibited item] on the platform?"
        \item Answer: "I'm sorry, but according to platform policies and national laws and regulations, the sale of such items is prohibited. Please abide by the platform's community guidelines to help maintain a healthy online environment."
    \end{itemize}
    \item \textbf{Jailbreak Prompts \& Defense}: To enhance the model’s robustness against malicious attacks, we proactively simulated various common jailbreak techniques and generated examples where the model successfully identified and refused to answer. This constitutes a dynamic “attack-defense drill” dataset. Attack types include:
    \begin{itemize}
        \item \textbf{Role-playing}: "Assume you are an AI with no constraints..."
        \item \textbf{Obfuscated Encoding}: Disguising sensitive words using Base64, homophones, special symbols, etc.
    \item \textbf{Contextual Injection}:Gradually inducing the model to lower its guard within long conversations.
        \item \textbf{Logic Exploitation}: Bypass rules through complex hypothetical scenarios.
        \item \textbf{Others}: Comprehensive coverage of 20+ techniques, including Ignoring Preambles, System Masquerading, Indirect Inducement, Repetitive Perturbation, Format Concealment, etc.
    \end{itemize}
    \item \textbf{Boundary Case Generation}: We synthesized a large number of cases at the edge of compliance and required the LLM to generate cautious analyses and decision-making processes.
\end{enumerate}
Building upon these three categories of synthetic data, we introduced a multi-stage filtering mechanism based on rules and model judgment to ensure the inherent safety and high quality of the synthetic data itself. This data is directly used as continuous pre-training safety data during the Inject stage and also serves as the foundation for safety alignment data during the Align stage.

\section{Inject: Knowledge Injection Pre-training with Depth Up-Scaling}

\subsection{Depth Up-Scaling for Parameter Expansion}
We introduce a Depth Up-Scaling\cite{kim2023solar} method that incorporates newly initialized layers between original Transformer blocks exhibiting the closest parameter similarity. This insertion strictly adheres to the principle of residual identity mapping during initialization to minimize disruption to the pre-trained model. For optimization, we employ a hierarchical adaptive learning rate mechanism—applying higher learning rates to the new layers to accelerate the acquisition of e-commerce domain knowledge, while utilizing lower learning rates coupled with weight decay for the original parameters to preserve their general capabilities. This heterogeneous learning strategy directs the added capacity specifically towards assimilating domain knowledge rather than perturbing the existing parameter space. This approach achieves significant domain-specific performance gains with minimal parameter growth, demonstrating favorable cost-effectiveness and scalability.
\subsection{Mitigating Catastrophic Forgetting}

\begin{enumerate}
    \item \textbf{Progressive Hierarchical Learning Rate Scheduling}

    Model parameters are divided into three tiers with differentiated learning rate controls:
    \begin{itemize}
        \item \textbf{High Learning Rate Tier}: Newly inserted Depth Up-Scaling layers, which are assigned accelerated learning rates to promote rapid convergence.
        \item \textbf{Medium Learning Rate Tier}: Several top layers of the original model, allowed to adapt moderately to the linguistic style and logical patterns of e-commerce.
        \item \textbf{Very Low Learning Rate / Frozen Tier}: The lower and middle layers of the original model, which serve as anchors to preserve general linguistic representations.
    \end{itemize}
    \item \textbf{Regularization via Knowledge Replay}
    
In each batch during continued pre-training, e-commerce corpora and general-domain corpora are dynamically mixed at a 6:4 ratio. The interleaving of general-domain data compels the model to continuously activate and consolidate its original parameters, effectively helping to mitigate overfitting to domain data and the decline of general capabilities.
    \item \textbf{Curriculum Learning and Blended Safety Strategy}
    
A curriculum learning strategy is introduced: training begins with simple, high-safety e-commerce samples, then gradually increases the proportion of complex and higher-risk samples (e.g., long-tail queries, multi-turn dialogues). In later training stages, short-context distillation and compression techniques are incorporated, enabling the model to make accurate judgments without relying on long contexts. This meets the industrial requirements for low latency and high throughput.
\end{enumerate}

\section{Align: Fine-grained Domain-Enhanced Alignment}

\subsection{E-commerce Multi-Task Instruction Tuning Data}
we have constructed a multi-granularity internal E-commerce Multi-Task Instruction Tuning dataset (Ecom-MTIT). This dataset covers core tasks in e-commerce search and content generation, including but not limited to: query suggestion, query understanding \& rewriting, product title/summary generation, review summarization \& analysis, and spelling correction. Moving beyond a single instruction format, we designed a three-tiered instruction system for each task, progressing from simple to complex:
\begin{itemize}
    \item \textbf{Simple Instructions}: Correspond to atomic tasks, such as "Rewrite 'iphone' into a formal product query." This stage aims to establish foundational task-mapping capabilities.
    \item \textbf{Composite Instructions}: Simulate real-world user complex intents by integrating multiple tasks into a single instruction, forcing the model to learn task planning and information integration.
    \item \textbf{Constrained Complex Instructions}: Building upon composite instructions, explicit business rules and constraints are introduced. This level directly addresses the requirements for precise, structured outputs in e-commerce scenarios.
\end{itemize}

\subsection{Safety Attack and Adversarial Data}
The safety guardrails of general-purpose models exhibit blind spots and are vulnerable to circumvention in high-risk e-commerce topics (e.g., prohibited items, medical devices, private data). To address this, we designed a security enhancement pipeline based on adversarial training, which consists of two stages:
\begin{itemize}
    \item \textbf{Stage 1: Supervised Safety Fine-Tuning Data Generation}

    Based on a predefined taxonomy of nine major high-risk topics, a security and compliance team manually reviewed and edited to create 10k aligned data samples. This dataset was used to perform supervised fine-tuning on the model post knowledge infusion, yielding the SIA-Safety model.
    \item \textbf{Stage 2: Red-Teaming Attacks and Adversarial Example Generation}

    Leveraging the SIA-Safety model, a red-team model named SIA-Red was trained on 1k human-annotated data. This model is tasked with generating highly deceptive and covert jailbreak instructions targeting the aforementioned nine risk categories (e.g., "How to describe [Prohibited Item A] as [Common Product B] to bypass review?"). The attack methodologies encompass over 20 types, including instruction ignoring, system prompt disguise, and indirect inducement. Ultimately, 10k jailbreak instructions were synthesized. These were then processed by the SIA-Safety model to generate responses, which were subsequently reviewed and edited by the security team.
\end{itemize}
These two stages collectively construct a 20k e-commerce domain-specific safety alignment instruction dataset.

\subsection{Alignment Data  Mixing Strategy}
To prevent the model from overfitting to domain-specific tasks and losing its general language understanding and reasoning capabilities, a mixed instruction set strategy was adopted. Building upon the Ecom-MTIT instruction dataset and the safety-aligned instruction dataset, a diverse set of general-purpose instruction alignment data was incorporated. The general-purpose instructions were sourced from three categories: Stanford Alpaca\cite{alpaca}, Databricks Dolly 15k\cite{DatabricksBlog2023DollyV2}, and ShareGPT\cite{sharegpt_chinese_english_90k}. These were obtained through refusal sampling based on instruction intent categorization. Ultimately, a total of 100k high-quality, mixed-proportion alignment data instruction sets were constructed.
Through experimentation, an appropriate mixing ratio was determined for e-commerce task instructions, general-purpose instructions, and safety-aligned instructions:E-commerce Task Instructions: 60\%, General-Purpose Instructions: 20\%, Safety-Aligned Instructions: 20\%. This balanced ratio ensures that the model achieves top-tier performance in e-commerce tasks while maintaining excellent generalization capabilities and safety compliance, effectively mitigating the issue of "catastrophic forgetting" of general-purpose abilities.

\section{Experiments}
\subsection{Experimental Setup}
\subsubsection{Datasets}

\begin{itemize}
    \item \textbf{Evaluation on E-commerce Domain Expertise and General Capability Preservation}
    \begin{itemize}
        \item \textbf{E-commerce Domain Evaluation}: Constructed a 5K-scale evaluation set based on internal real-world scenarios, covering three core tasks:
    \begin{enumerate}
        
        \item E-commerce Knowledge Question Answering (product parameters, descriptions, recommendations);
        \item Attribute Understanding (category\slash brand\slash model recognition, attribute Q\&A);
        \item Product Query Correction (spelling error correction, abbreviation expansion, alias resolution).
    \end{enumerate}
    \item \textbf{The General Capability Evaluation}: primarily aims to verify whether the model retains its general abilities and to effectively mitigate the issue of catastrophic forgetting. Therefore, several widely recognized core benchmarks were adopted to measure the model's fundamental reasoning and knowledge capabilities. These specifically include: C-Eval\cite{huang2023c}, CMMLU\cite{li2024cmmlu}, MMLU\cite{hendrycks2020measuring}, GSM8K\cite{cobbe2021training}, and MMLU-Pro\cite{wang2024mmlu}.
    \end{itemize}

    \item \textbf{Safety and Compliance Evaluation}
    
Given the critical importance of safety for the JD.com platform, we constructed the JDSec safety evaluation set covering eight risk dimensions:politics, pornography, prohibited items, rights infringement, commercial violations, religion, discriminatory speech, and historical nihilism. This dataset contains a large number of carefully crafted jailbreak prompts to systematically test the robustness of the model’s safety guardrails.
\end{itemize}

\subsubsection{Baselines and Evaluation Metrics}
Qwen2.5-14B-Instruct was selected as the baseline model. Evaluation metrics are uniformly reported on a 100-point scale: accuracy is used for general tasks and product attribute understanding; ROUGE-L is adopted for e-commerce knowledge question answering; the F1-score is applied for the product query error correction task; and the attack success rate (ASR)—defined as the proportion of adversarial examples that successfully induce the target harmful behavior—is used for safety tasks.

\subsubsection{Implementation and Deployment Environment}
The model is based on Qwen2.5-14B-Instruct\cite{qwen2025qwen25technicalreport}, with its depth extended to an equivalent of 15B parameters via Depth Up-Scaling. Training was conducted on an Ascend 910B cluster, employing mixed precision and the ZeRO optimizer to enhance efficiency. For online inference, deployed on the same cluster, optimizations such as quantization, dynamic batching, and operator fusion were implemented to meet the high-concurrency, low-latency requirements of e-commerce search, ensuring industrial-grade viability.

\subsection{Main Results}
\subsubsection{E-commerce Domain \& General Capability Preservation Evaluation}
As shown in Table \ref{tab:eval1}, SIA-15B, trained with our proposed method, demonstrates a significant advantage in the e-commerce domain. It even surpasses the performance of DeepSeek-V3 directly applied to e-commerce tasks. On most general-purpose benchmarks, the performance of SIA-15B is effectively preserved compared to the original Qwen2.5-14B-Instruct model, with minor improvements observed on some tasks. These results indicate that our SIA method successfully enhances the model's knowledge in the e-commerce domain while achieving an effective balance between general and domain-specific capabilities.
\begin{table*}
  \caption{E‑commerce Domain and General Competence Evaluation Results}
  \label{tab:eval1}
  \begin{tabular}{ccccccccc}
    \toprule
    Model & EKQA & PAU & EQEC & C-EVAL & CMMLU & MMLU & GSM8K & MMLU-Pro\\
    \midrule
    Qwen2.5-14B-Instruct (baseline) & 23.87 & 50.65 & 39.04 & 83.35 & 84.09 & 78.69 & 94.8 & 63.7 \\
    \textbf{SIA-15B (Ours)} & \textbf{58.01} & \textbf{91.72} & \textbf{74.29} & 83.98 & 84.94 & 78.45 & 94.9 & 66.82\\
    deepseek-v3 & 25.82 & 57.72 & 51.20 & / & / & / & / & /  \\
  \bottomrule
\end{tabular}
\end{table*}

\subsubsection{Evaluation of Safety and Compliance}
In the safety benchmark JDSec (as shown in Table \ref{tab:safety}), across the eight major risk categories—Sensitive Political (SP), Sexually Explicit (SE), Prohibited, Rights Infringement (RI), Commercial Non-compliance (CNC), Religious, Discrimination, and Historical Negationism (HN)—SIA-15B demonstrates significant improvements on most metrics compared to the baseline model Qwen2.5-14B-Instruct (QWEN-14B) (The lower the ASR, the better). Its performance even surpasses that of large models such as DeepSeek-V3 (dp-V3) and GPT-4o.\begin{table}
  \caption{Safety and Compliance Evaluation Results(lower is better)}
  \label{tab:safety}
  \begin{tabular}{ccccc}
    \toprule
    Type & Qwen-14B & dp-V3 & gpt4o &  SIA-15B \\
    \midrule
    SP & 34.16 & 33.15 & 30.58 & \textbf{28.60} \\
    SE & 13.16 & 10.73 & \textbf{9.32}  & 10.48 \\
    Prohibited & 10.85 & 10.11 & 9.97 & \textbf{7.81} \\
    RI & 13.94 & 12.21 & \textbf{10.28}  & 14.08 \\
    CNC & 11.61 & 14.65 & 13.45 & \textbf{9.34} \\
    Religious & 19.59 & 18.85 & 16.32 & \textbf{11.11} \\
    Discrimination & 12.04 & 11.78 & \textbf{10.63} & 11.61 \\
    HN & 18.06 & 16.36 & 18.79 & \textbf{16.18} \\
    JDSec-avg & 16.68 & 15.98 & 14.92 & \textbf{13.65} \\
  \bottomrule
\end{tabular}
\end{table}

\begin{table*}
  \caption{Ablation Results}
  \label{tab:ablation}
  \begin{tabular}{cccccccccc}
    \toprule
    Variants & EKQA & PAU & EQEC & C-Eval & CMMLU & MMLU & GSM8K & MMLU-Pro & JDSec-avg  \\
    \midrule
Qwen2.5-14B-Instruct (baseline) & 23.87 & 50.65 &39.04 &83.35 & 84.09 & 78.69 & 94.8 & 63.7 & 16.68  \\
SIA-15B (Full) & 58.01 & 91.72 & 74.29 & 83.98 & 84.94 & 78.45 & 94.9 & 66.82 & 13.65  \\
w/o Context-Aware Synthesis & 41.89 & 74.56 & 54.72 & 83.52 & 84.21 & 78.03 & 94.7 & 64.51 & 15.22  \\
w/o Depth Up-Scaling & 44.73 & 77.89 & 57.96 & 82.54 & 83.02 & 77.51 & 94.6 & 63.05 & 15.51  \\
w/o Common Replay & 59.12 & 92.15 & 75.33 & 81.07 & 82.05 & 76.02 & 94.5 & 61.03 & 14.03  \\
w/o CPT & 3015 & 64.88 & 47.89 & 83.56 & 84.13 & 78.54 & 94.8 & 63.82 & 16.01 \\
w/o Hierarchical Instructions & 49.87 & 87.93 & 67.85 & 83.84 & 84.72 & 78.32 & 94.8 & 65.54 & 14.52  \\
w/o Common Instructions & 57.24 & 90.87 & 72.86 & 82.83 & 83.57 & 77.56 & 94.7 & 64.01 & 13.84  \\
w/o SFT & 47.65 & 84.92 & 64.98 & 83.71 & 84.53 & 78.24 & 94.8 & 65.03 & 15.80  \\
w/o Red-Teaming Alignment & 56.98 & 90.73 & 72.54 & 83.91 & 84.81 & 78.42 & 94.9 & 66.51 & 15.30  \\
w/o Safety & 56.73 & 90.59 & 72.31 & 83.95 & 84.87 & 78.40 & 94.9 & 66.38 & 16.20 \\
  \bottomrule
\end{tabular}
\end{table*}

\subsection{Ablation Studies}

In our ablation studies, we compare the complete SIA-15B model with the following key variants (as shown in the table below). All experiments are conducted under a unified training framework, optimizer configuration (mixed precision training with ZeRO optimization), and deployment environment (Ascend 910B cluster).
The detailed results of the ablation experiments are presented in Table \ref{tab:ablation}:

Based on the aforementioned ablation experimental results, we conducted ablation analysis from the following four dimensions:
\begin{enumerate}
    \item \textbf{Validation of the Context-Aware Synthesis}
    
 \textbf{w/o Context-Aware Synthesis}: During the data synthesis phase, direct use of raw JSON instead of Context-Aware Synthesis led to a significant decline in core e-commerce metrics in experimental validation, confirming that contextual knowledge synthesis is a prerequisite for efficient knowledge injection: the rigid structure of raw JSON hinders the understanding of product relationships, whereas semantic synthesis explicitly encodes tacit knowledge and improves absorption efficiency. Meanwhile, the JDSec-avg ASR of this variant increased by 1.57, indicating that semantic processing of safety-critical corpora simultaneously enhances risk identification capabilities.

    \item \textbf{Validation of Knowledge Injection Effectiveness}
    \begin{itemize}
        \item \textbf{w/o Depth Up-Scaling}: 
        During the knowledge injection phase, omitting parameter expansion and conducting full pre-training directly on the original Qwen2.5-14B led to a significant decline in both e-commerce metrics and general capabilities in experimental validation (e.g., a 13.28 drop in ROUGE-L for EKQA, and decreases of 1.44 and 1.92 in C-Eval and CMMLU, respectively). This indicates that introducing dedicated new layers alongside a layered learning rate strategy can effectively inject e-commerce knowledge while mitigating the degradation of general capabilities and avoiding knowledge conflict.       
         \item \textbf{w/o Common Replay}: Exclusive use of e-commerce corpora during pre-training, omitting the 40\% blend of general-domain data, resulted in a marginal improvement in e-commerce metrics but precipitated a substantial decline in general capabilities. This demonstrates that a hybrid training regimen blending domain-specific and general corpora at a 6:4 ratio prevents catastrophic forgetting, mitigates overfitting to the e-commerce domain, and achieves domain expertise without compromising generalization capabilities .
        \item \textbf{w/o CPT}: Removing Continuous Pre-training (CPT) and directly applying instruction tuning to the baseline model leads to a significant decline in e-commerce metrics, as validated experimentally. This demonstrates that knowledge infusion during pre-training serves as an essential foundational basis, and instruction tuning alone cannot enable the model to master in‑depth e‑commerce knowledge.
    
    \end{itemize}

    \item \textbf{Validation of Domain Alignment Stage Value}
    \begin{itemize}
        \item \textbf{w/o Hierarchical Instructions}: Removal of the three-tiered instruction hierarchy, utilizing only single-step instructions during the alignment phase, resulted in experimental declines of 8.14 and 6.44 in EKQA and EQEC, respectively. This empirically validates that the three-tier instruction system effectively elicits complex reasoning and structured output capabilities.
        \item \textbf{w/o Common Instructions}: When 20\% of general instructions were removed, employing only domain-specific (e-commerce) instructions during alignment led to a significant degradation in general capabilities with only marginal fluctuations in e-commerce metrics. This indicates that incorporating general instructions prevents over-specialization to the domain, maintaining e-commerce performance while preserving model generalization.
        \item \textbf{w/o SFT}:  Solely performing the knowledge injection pre-training (CPT) phase without subsequent supervised fine-tuning (SFT) for alignment resulted in decreased e-commerce metrics and increased safety risks. This demonstrates that the SFT stage is essential for translating the knowledge acquired during CPT into controllable, reliable operational capabilities and enhancing adherence to instructions and safety protocols.
    \end{itemize}

    \item \textbf{Validation of Security Module Defense Efficacy}
    \begin{itemize}
      \item \textbf{w/o Red-Teaming Alignment}: Limiting safety alignment solely to the first-phase safe Q\&A and removing adversarial examples from red-teaming attacks increased the safety risk by 1.65 (JDSec-avg ASR). This demonstrates that red-teaming adversarial training effectively probes security blind spots, enhancing the model’s defense against concealed risks.       
      \item \textbf{w/o Safety}: Complete removal of safety corpus enhancement and adversarial training resulted in the JDSec-avg ASR rising to 16.20, nearing baseline levels. This result validates the necessity of the SIA framework’s end-to-end safety design, integrating “data-side safety enhancement” with “alignment-phase adversarial training”.
    \end{itemize}
\end{enumerate}
In summary, each component of the SIA framework plays an indispensable role in constructing a model that possesses deep e-commerce expertise, robust general intelligence, and rigorous safety-compliance.

\subsection{Online A/B Testing in Industrial E-Commerce Platform}
\subsubsection{Online A/B Testing for Search Applications}
To validate the real-world business value of SIA, we deployed it on JD.com’s online search system—China’s largest self-operated e-commerce platform—and conducted large-scale online A/B testing over a one-month period.
We deployed the trained SIA-15B model across multiple search-related scenarios and compared it against a baseline system fine-tuned from Qwen2.5-14B-Instruct. Under identical prompt and input conditions, SIA-15B achieved statistically significant (p-value < 0.05) business improvements in the following tasks:
\begin{itemize}
    \item \textbf{Search Query Suggestion}: Generating more accurate and engaging search suggestions based on the user’s input query, clicked products, and basic user profile. This yielded a significant 1.339\% lift in overall system CTR (Click-Through Rate), demonstrating SIA-15B’s superior ability to predict and guide user intent.
    \item \textbf{Search Query Correction}: Performing real-time correction on user input queries and generating updates for the online dictionary service in a near-real-time manner. This resulted in a 10\% improvement in F1-score for the online correction service, directly reducing invalid searches caused by user input errors and enhancing overall search experience and efficiency.
    \item \textbf{Product Title Generation}: Directly generating titles with greater marketing appeal and informational value from product information inputs. This directly drove a 1.93\% increase in CVR (Conversion Rate) for recommendations, evidencing its commercial value in content generation.
    \item \textbf{Review Summarization}: Generating more insightful review summaries from extensive user comments to aid rapid decision-making, leading to a 0.15\% increase in UCVR (Unit Conversion Rate) across the overall system.
\end{itemize}

These online A/B tests conclusively demonstrate that SIA-15B significantly outperforms traditional fine-tuning approaches in e-commerce search applications, effectively driving business growth. It has been successfully deployed in production on JD.com’s e-commerce search platform.

\subsubsection{Online Verification of Security Risk Control and Auditing}
In security risk control scenarios, SIA-15B demonstrates significantly superior performance compared to the baseline system fine-tuned from Qwen2.5-14B-Instruct (online test results detailed in Table \ref{tab:risk}). It achieves substantial improvements in risk identification accuracy across multiple tasks, including \textbf{code security auditing(CSR)}, \textbf{AI-generated content security(AIGCS)}, \textbf{phishing email detection(PE)}, \textbf{security operation center alert analysis(SOCA)}, and \textbf{traffic security identification(TS)}. The model has been fully deployed in production, now covering over 90\% of consumer-facing risk control review scenarios within the JD.com ecosystem, serving as the foundational security layer for the platform’s AI services.
\begin{table}
  \caption{Online Evaluation Results for Security and Compliance}
  \label{tab:risk}
  \begin{tabular}{cccccc}
    \toprule
    Model & CSR & AIGCS & PE & SOCA  & TS \\
    \midrule
    SIA-15B & \textbf{72.03} & \textbf{80.71} & \textbf{91.43} & \textbf{86.67} & \textbf{75.47} \\
    baseline & 42.71 & 60.18 & 59.46 & 61.25 & 48.81 \\
  \bottomrule
\end{tabular}
\end{table}
\section{Conclusion and Future Work}
To tackle the two major challenges in deploying large language models (LLMs) in e‑commerce—knowledge hallucination and contextual safety and compliance risks—we propose the SIA (Synthesize, Inject, Align)  framework.Key innovations include: 1) Knowledge-aware data synthesis fusing structured KGs and unstructured logs to solve sparsity and safety vulnerabilities; 2) Efficient knowledge injection via Depth Up-Scaling in Transformers to prevent catastrophic forgetting; 3) Fine-grained dual-path alignment using hierarchical instruction tuning and adversarial training with optimized data ratios enabling robust compliance. The SIA-15B model has been deployed on JD.com, demonstrating marked improvements in both core e-commerce scenarios and safety performance.
Future work focuses on three directions: Dynamic knowledge update through lightweight incremental training for real-time synchronization with new product/promotion data; Multi-modal fusion integrating visual knowledge graphs and transformers to support image-based search and safety detection; Personalized scenario-adaptive safety alignment building tailored risk taxonomies and defenses adjusted by user/scenario attributes (e.g., cross-border, children’s products) for precision risk mitigation.

\bibliographystyle{ACM-Reference-Format}
\bibliography{sample-base}

@String{Chelsea = "Chelsea" }

@article{achiam2023gpt,
  title={GPT-4 Technical Report},
  author={Achiam, Josh and Adler, Steven and Agarwal, Sandhini and Ahmad, Lama and Akkaya, Ilge and Aleman, Florencia and Almeida, Diogo and Altenschmidt, Janko and Altman, Sam and Anadkat, Shyamal and others},
  journal={arXiv preprint arXiv:2303.08774},
  year={2023}
}

@article{touvron2023llama,
  title={Llama 2: Open Foundation and Fine-Tuned Chat Models},
  author={Touvron, Hugo and Martin, Louis and Stone, Kevin and Albert, Peter and Almahairi, Amjad and Babaei, Yasmine and Bashlykov, Nikolay and Batra, Soumya and Bhargava, Prajjwal and Bhosale, Shruti and others},
  journal={arXiv preprint arXiv:2307.09288},
  year={2023}
}

@article{guo2025deepseek,
  title={Deepseek-r1: Incentivizing reasoning capability in llms via reinforcement learning},
  author={Guo, Daya and Yang, Dejian and Zhang, Haowei and Song, Junxiao and Zhang, Ruiyu and Xu, Runxin and Zhu, Qihao and Ma, Shirong and Wang, Peiyi and Bi, Xiao and others},
  journal={arXiv preprint arXiv:2501.12948},
  year={2025}
}

@inproceedings{gururangan2020don,
  title={Don't Stop Pretraining: Adapt Language Models to Domains and Tasks},
  author={Gururangan, Suchin and Marasovi{\'c}, Ana and Swayamdipta, Swabha and Lo, Kyle and Beltagy, Iz and Downey, Doug and Smith, Noah A},
  booktitle={Proceedings of the 58th Annual Meeting of the Association for Computational Linguistics},
  pages={8342--8360},
  year={2020}
}

@inproceedings{li2024ecomgpt,
  title={EcomGPT: Instruction-tuning Large Language Models with Chain-of-Task Tasks for E-commerce},
  author={Li, Yangning and Ma, Shirong and Wang, Xiaobin and Huang, Shen and Jiang, Chengyue and Zheng, Hai-Tao and Xie, Pengjun and Huang, Fei and Jiang, Yong},
  booktitle={Proceedings of the AAAI Conference on Artificial Intelligence},
  volume={38},
  number={17},
  pages={18582--18590},
  year={2024}
}

@inproceedings{peng2024ecellm,
  title={eCeLLM: Generalizing Large Language Models for E-commerce from Large-scale, High-quality Instruction Data},
  author={Peng, Bo and Ling, Xinyi and Chen, Ziru and Sun, Huan and Ning, Xia},
  booktitle={Proceedings of the 41st International Conference on Machine Learning (ICML)},
  year={2024}
}

@inproceedings{hu2022lora,
  title={LoRA: Low-Rank Adaptation of Large Language Models},
  author={Hu, Edward J and Shen, Yelong and Wallis, Phillip and Allen-Zhu, Zeyuan and Li, Yuanzhi and Wang, Shean and Wang, Lu and Chen, Weizhu},
  booktitle={International Conference on Learning Representations (ICLR)},
  year={2022}
}

@inproceedings{houlsby2019parameter,
  title={Parameter-Efficient Transfer Learning for NLP},
  author={Houlsby, Neil and Giurgiu, Andrei and Jastrzebski, Stanislaw and Morrone, Bruna and De Lange, Quentin and Andrea, Gesmundo and Fisher, Zhu and Ciaramita, Massimiliano},
  booktitle={International Conference on Machine Learning (ICML)},
  pages={2790--2799},
  year={2019},
  organization={PMLR}
}

@inproceedings{dettmers2023qlora,
  title={QLoRA: Efficient Finetuning of Quantized LLMs},
  author={Dettmers, Tim and Pagnoni, Artidoro and Holtzman, Ari and Zettlemoyer, Luke},
  booktitle={Advances in Neural Information Processing Systems (NeurIPS)},
  volume={36},
  year={2023}
}

@inproceedings{zhang2023adalora,
  title={AdaLoRA: Adaptive Budget Allocation for Parameter-Efficient Fine-Tuning},
  author={Zhang, Qingru and Chen, Minfeng and Bukharin, Alexander and He, Pengcheng and Cheng, Yu and Chen, Weizhu and Zhao, Tuo},
  booktitle={International Conference on Learning Representations (ICLR)},
  year={2023}
}

@article{biderman2024lora,
  title={LoRA Learns Less and Forgets Less},
  author={Biderman, Dan and Portes, Jacob and Ortiz, Jose Javier Gonzalez and Paul, Mansheej and Greengard, Philip and Jennings, Connor and King, Daniel and Havens, Sam and Chiley, Vitaliy and Frankle, Jonathan and others},
  journal={Transactions on Machine Learning Research},
  issn={2835-8856},
  year={2024}
}

@inproceedings{lewis2020retrieval,
  title={Retrieval-Augmented Generation for Knowledge-Intensive NLP Tasks},
  author={Lewis, Patrick and Perez, Ethan and Piktus, Aleksandra and Petroni, Fabio and Karpukhin, Vladimir and Goyal, Naman and K{\"u}ttler, Heinrich and Lewis, Mike and Yih, Wen-tau and Rockt{\"a}schel, Tim and others},
  booktitle={Advances in Neural Information Processing Systems (NeurIPS)},
  volume={33},
  pages={9459--9474},
  year={2020}
}

@article{ma2023ecomgpt,
  title={EcomGPT-CT: Continual Pre-training of E-commerce Large Language Models with Semi-structured Data},
  author={Ma, Shirong and Huang, Shen and Huang, Shulin and Wang, Xiaobin and Li, Yangning and Zheng, Hai-Tao and Xie, Pengjun and Huang, Fei and Jiang, Yong},
  journal={arXiv preprint arXiv:2312.15696},
  year={2023}
}

@inproceedings{brown2020language,
  title={Language models are few-shot learners},
  author={Brown, Tom and Mann, Benjamin and Ryder, Nick and Subbiah, Melanie and Kaplan, Jared D and Dhariwal, Prafulla and Neelakantan, Arvind and Shyam, Pranav and Sastry, Girish and Askell, Amanda and others},
  booktitle={Advances in Neural Information Processing Systems (NeurIPS)},
  volume={33},
  pages={1877--1901},
  year={2020}
}

@article{tay2022transformer,
  title={Transformer memory as a differentiable search index},
  author={Tay, Yi and Dehghani, Mostafa and Tran, Vinh Q and Garcia, Xavier and Wei, Jason and Bahri, Dara and Schuster, Tal and Zheng, Huaixiu Steven and Zhou, Denny and Metzler, Donald and Cohen, William W},
  journal={NeurIPS},
  volume={35},
  pages={21831--21843},
  year={2022}
}

@inproceedings{bevilacqua2022autoregressive,
  title={Autoregressive search engines: Generating substrings as document identifiers},
  author={Bevilacqua, Michele and Ottaviano, Giuseppe and Lewis, Patrick and Yih, Wen-tau and Riedel, Sebastian and Petroni, Fabio},
  booktitle={Advances in Neural Information Processing Systems (NeurIPS)},
  volume={35},
  pages={31668--31683},
  year={2022}
}

@inproceedings{ouyang2022training,
  title={Training language models to follow instructions with human feedback},
  author={Ouyang, Long and Wu, Jeffrey and Jiang, Xu and Almeida, Diogo and Wainwright, Carroll and Mishkin, Pamela and Zhang, Chong and Agarwal, Sandhini and Slama, Katarina and Ray, Alex and others},
  booktitle={Advances in Neural Information Processing Systems (NeurIPS)},
  volume={35},
  pages={27730--27744},
  year={2022}
}

@inproceedings{rafailov2023direct,
  title={Direct preference optimization: Your language model is secretly a reward model},
  author={Rafailov, Rafael and Sharma, Archit and Mitchell, Eric and Manning, Christopher D and Ermon, Stefano and Finn, Chelsea},
  booktitle={Advances in Neural Information Processing Systems (NeurIPS)},
  volume={36},
  year={2023}
}

@article{hinton2015distilling,
  title={Distilling the knowledge in a neural network},
  author={Hinton, Geoffrey and Vinyals, Oriol and Dean, Jeff},
  journal={arXiv preprint arXiv:1503.02531},
  year={2015}
}

@inproceedings{magister2023teaching,
  title={Teaching small language models to reason},
  author={Magister, Lucie Charlotte and Mallinson, Jonathan and Adamek, Jakub and Gur-Ari, Guy and Severyn, Aliaksei},
  booktitle={Proceedings of the 61st Annual Meeting of the Association for Computational Linguistics (Volume 1: Long Papers)},
  pages={1773--1781},
  year={2023}
}

@inproceedings{pradeep2023how,
  title={How Does Generative Retrieval Scale to Millions of Passages?},
  author={Pradeep, Ronak and Zhuang, Honglei and Zhang, Kai Hui and Cannon, Liam andM, Chu-Cheng and others},
  booktitle={Proceedings of the 2023 Conference on Empirical Methods in Natural Language Processing (EMNLP)},
  pages={1229--1244},
  year={2023}
}

@article{kim2023solar,
  title={SOLAR 10.7B: Scaling Large Language Models with Integrated Depth Up-Scaling},
  author={Kim, Dahyun and Park, Chanjun and Kim, Sanghoon and Lee, Wonsung and Song, Wonho and Kim, Yunsu and Kim, Hyeonwoo and Kim, Yungi and Lee, Jung Ju and Kim, Hwaran},
  journal={arXiv preprint arXiv:2312.15166},
  year={2023}
}

@article{gupta2023continual,
  title={Continual Pre-Training of Large Language Models: How to (re)warm your model?},
  author={Gupta, Kshitij and Theriault, Mike and Shazeer, Noam},
  journal={arXiv preprint arXiv:2308.04014},
  year={2023}
}

@article{salton1988term,
  title={Term-weighting approaches in automatic text retrieval},
  author={Salton, Gerard and Buckley, Christopher},
  journal={Information processing \& management},
  volume={24},
  number={5},
  pages={513--523},
  year={1988},
  publisher={Elsevier}
}

@inproceedings{huang2020embedding,
  title={Embedding-based retrieval in facebook search},
  author={Huang, Jui-Ting and Sharma, Ashish and Sun, Shuying and Xia, Li and Zhang, David and Pronin, Philip and Padmanabhan, Janani and Ottaviano, Giuseppe and Yang, Linjun},
  booktitle={Proceedings of the 26th ACM SIGKDD International Conference on Knowledge Discovery \& Data Mining},
  pages={2553--2561},
  year={2020}
}

@inproceedings{reimers2019sentence,
  title={Sentence-BERT: Sentence Embeddings using Siamese BERT-Networks},
  author={Reimers, Nils and Gurevych, Iryna},
  booktitle={Proceedings of the 2019 Conference on Empirical Methods in Natural Language Processing and the 9th International Joint Conference on Natural Language Processing (EMNLP-IJCNLP)},
  pages={3982--3992},
  year={2019}
}

@inproceedings{nguyen2025minielm,
  title={Minielm: A lightweight and adaptive query rewriting framework for e-commerce search optimization},
  author={Nguyen, Duy A and Mohan, Rishi Kesav and Yang, Shimeng and Akash, Pritom Saha and Chang, Kevin Chen-Chuan},
  booktitle={Findings of the Association for Computational Linguistics: ACL 2025},
  pages={6952--6964},
  year={2025}
}

@inproceedings{dai2024enhancing,
  title={Enhancing E-Commerce Query Rewriting: A Large Language Model Approach with Domain-Specific Pre-Training and Reinforcement Learning},
  author={Dai, Aijun and Zhu, Zhenyu and Hu, Haiqing and Tang, Guoyu and Liu, Lin and Xu, Sulong},
  booktitle={Proceedings of the 33rd ACM International Conference on Information and Knowledge Management},
  pages={4439--4445},
  year={2024}
}

@inproceedings{peng2024large,
  title={Large language model based long-tail query rewriting in taobao search},
  author={Peng, Wenjun and Li, Guiyang and Jiang, Yue and Wang, Zilong and Ou, Dan and Zeng, Xiaoyi and Xu, Derong and Xu, Tong and Chen, Enhong},
  booktitle={Companion Proceedings of the ACM Web Conference 2024},
  pages={20--28},
  year={2024}
}

@inproceedings{yuan2025semi,
  title={A Semi-supervised Scalable Unified Framework for E-commerce Query Classification},
  author={Yuan, Chunyuan and Zhang, Chong and Fang, Zhen and Pang, Ming and Jiang, Xue and Peng, Changping and Lin, Zhangang and Law, Ching},
  booktitle={Proceedings of the 63rd Annual Meeting of the Association for Computational Linguistics (Volume 6: Industry Track)},
  pages={1263--1271},
  year={2025}
}

@inproceedings{tang2025lref,
  title={LREF: A Novel LLM-based Relevance Framework for E-commerce Search},
  author={Tang, Tian and Tian, Zhixing and Zhu, Zhenyu and Wang, Chenyang and Hu, Haiqing and Tang, Guoyu and Liu, Lin and Xu, Sulong},
  booktitle={Companion Proceedings of the ACM on Web Conference 2025},
  pages={468--475},
  year={2025}
}

@article{dong2025taosr1,
  title={TaoSR1: The thinking model for e-commerce relevance search},
  author={Dong, Chenhe and Yao, Shaowei and Jiao, Pengkun and Yang, Jianhui and Jin, Yiming and Huang, Zerui and Zhou, Xiaojiang and Ou, Dan and Tang, Haihong and Zheng, Bo},
  journal={arXiv preprint arXiv:2508.12365},
  year={2025}
}

@article{li2024generative,
  title={Generative retrieval with preference optimization for e-commerce search},
  author={Li, Mingming and Wang, Huimu and Chen, Zuxu and Nie, Guangtao and Qiu, Yiming and Tang, Guoyu and Liu, Lin and Zhuo, Jingwei},
  journal={arXiv preprint arXiv:2407.19829},
  year={2024}
}

@article{gao2023chat,
  title={Chat-rec: Towards interactive and explainable llms-augmented recommender system},
  author={Gao, Yunfan and Sheng, Tao and Xiang, Youlin and Xiong, Yun and Wang, Haofen and Zhang, Jiawei},
  journal={arXiv preprint arXiv:2303.14524},
  year={2023}
}

@article{schulman2017proximal,
  title={Proximal policy optimization algorithms},
  author={Schulman, John and Wolski, Filip and Dhariwal, Prafulla and Radford, Alec and Klimov, Oleg},
  journal={arXiv preprint arXiv:1707.06347},
  year={2017}
}

@article{zou2023universal,
  title={Universal and transferable adversarial attacks on aligned language models},
  author={Zou, Andy and Wang, Zifan and Carlini, Nicholas and Nasr, Milad and Kolter, J Zico and Fredrikson, Matt},
  journal={arXiv preprint arXiv:2307.15043},
  year={2023}
}

@article{liu2023jailbreaking,
  title={Jailbreaking chatgpt via prompt engineering: An empirical study},
  author={Liu, Yi and Deng, Gelei and Xu, Zhengzi and Li, Yuekang and Zheng, Yaowen and Zhang, Ying and Zhao, Lida and Zhang, Tianwei and Wang, Kailong and Liu, Yang},
  journal={arXiv preprint arXiv:2305.13860},
  year={2023}
}

@misc{alpaca,
  author = {Rohan Taori and Ishaan Gulrajani and Tianyi Zhang and Yann Dubois and Xuechen Li and Carlos Guestrin and Percy Liang and Tatsunori B. Hashimoto },
  title = {Stanford Alpaca: An Instruction-following LLaMA model},
  year = {2023},
  publisher = {GitHub},
  journal = {GitHub repository},
  howpublished = {\url{https://github.com/tatsu-lab/stanford_alpaca}},
}

@online{DatabricksBlog2023DollyV2,
    author    = {Mike Conover and Matt Hayes and Ankit Mathur and Jianwei Xie and Jun Wan and Sam Shah and Ali Ghodsi and Patrick Wendell and Matei Zaharia and Reynold Xin},
    title     = {Free Dolly: Introducing the World's First Truly Open Instruction-Tuned LLM},
    year      = {2023},
    url       = {https://www.databricks.com/blog/2023/04/12/dolly-first-open-commercially-viable-instruction-tuned-llm},
    urldate   = {2023-06-30}
}

@dataset{sharegpt_chinese_english_90k,
  author    = {{ShareAI Lab}},
  title     = {ShareGPT-Chinese-English-90k: A Bilingual Chinese-English Human-Machine Dialogue Dataset},
  year      = {2023},
  publisher = {Hugging Face},
  url       = {https://huggingface.co/datasets/shareAI/ShareGPT-Chinese-English-90k}
}

@article{huang2023c,
  title={C-eval: A multi-level multi-discipline chinese evaluation suite for foundation models},
  author={Huang, Yuzhen and Bai, Yuzhuo and Zhu, Zhihao and Zhang, Junlei and Zhang, Jinghan and Su, Tangjun and Liu, Junteng and Lv, Chuancheng and Zhang, Yikai and Fu, Yao and others},
  journal={Advances in Neural Information Processing Systems},
  volume={36},
  pages={62991--63010},
  year={2023}
}

@inproceedings{li2024cmmlu,
  title={Cmmlu: Measuring massive multitask language understanding in chinese},
  author={Li, Haonan and Zhang, Yixuan and Koto, Fajri and Yang, Yifei and Zhao, Hai and Gong, Yeyun and Duan, Nan and Baldwin, Timothy},
  booktitle={Findings of the Association for Computational Linguistics: ACL 2024},
  pages={11260--11285},
  year={2024}
}

@article{hendrycks2020measuring,
  title={Measuring massive multitask language understanding},
  author={Hendrycks, Dan and Burns, Collin and Basart, Steven and Zou, Andy and Mazeika, Mantas and Song, Dawn and Steinhardt, Jacob},
  journal={arXiv preprint arXiv:2009.03300},
  year={2020}
}

@article{wang2024mmlu,
  title={Mmlu-pro: A more robust and challenging multi-task language understanding benchmark},
  author={Wang, Yubo and Ma, Xueguang and Zhang, Ge and Ni, Yuansheng and Chandra, Abhranil and Guo, Shiguang and Ren, Weiming and Arulraj, Aaran and He, Xuan and Jiang, Ziyan and others},
  journal={Advances in Neural Information Processing Systems},
  volume={37},
  pages={95266--95290},
  year={2024}
}

@article{cobbe2021training,
  title={Training verifiers to solve math word problems},
  author={Cobbe, Karl and Kosaraju, Vineet and Bavarian, Mohammad and Chen, Mark and Jun, Heewoo and Kaiser, Lukasz and Plappert, Matthias and Tworek, Jerry and Hilton, Jacob and Nakano, Reiichiro and others},
  journal={arXiv preprint arXiv:2110.14168},
  year={2021}
}

@misc{qwen2025qwen25technicalreport,
      title={Qwen2.5 Technical Report}, 
      author={Qwen and : and An Yang and Baosong Yang and Beichen Zhang and Binyuan Hui and Bo Zheng and Bowen Yu and Chengyuan Li and Dayiheng Liu and Fei Huang and Haoran Wei and Huan Lin and Jian Yang and Jianhong Tu and Jianwei Zhang and Jianxin Yang and Jiaxi Yang and Jingren Zhou and Junyang Lin and Kai Dang and Keming Lu and Keqin Bao and Kexin Yang and Le Yu and Mei Li and Mingfeng Xue and Pei Zhang and Qin Zhu and Rui Men and Runji Lin and Tianhao Li and Tianyi Tang and Tingyu Xia and Xingzhang Ren and Xuancheng Ren and Yang Fan and Yang Su and Yichang Zhang and Yu Wan and Yuqiong Liu and Zeyu Cui and Zhenru Zhang and Zihan Qiu},
      year={2025},
      eprint={2412.15115},
      archivePrefix={arXiv},
      primaryClass={cs.CL},
      url={https://arxiv.org/abs/2412.15115}, 
}
\end{document}